\documentclass[a4paper,fleqn]{cas-sc}

\usepackage[authoryear]{natbib}

\usepackage{url} 
\usepackage{booktabs}
\usepackage{multirow}
\usepackage{soul}
\usepackage{color}
\usepackage{makecell}
\usepackage{colortbl}
\usepackage{pifont}
\def\ourmodel{\text{TSRNet}}   

\usepackage[linesnumbered,ruled,vlined]{algorithm2e}

\usepackage{algpseudocode}
\usepackage{threeparttable}
\usepackage{makecell}

\def\tsc#1{\csdef{#1}{\textsc{\lowercase{#1}}\xspace}}
\tsc{WGM}
\tsc{QE}

\begin{document}
\let\WriteBookmarks\relax
\def\floatpagepagefraction{1}
\def\textpagefraction{.001}

\shorttitle{Medical Image Analysis (2025)}    

\shortauthors{Yaolei Qi et~al.}  

\title [mode = title]{Rethinking the Detail-Preserved Completion of Complex Tubular Structures based on Point Cloud: a Dataset and a Benchmark}  

\tnotemark[1] 

\tnotetext[1]{This research was supported by the Postgraduate Research $\&$ Practice Innovation Program of Jiangsu Province, the Fundamental Research Funds for the Central Universities (KYCX22$\_$0239). We thank the Big Data Computing Center of Southeast University for providing the facility support.} 

%

\author[1]{Yaolei Qi}[type=editor,
                       orcid=0000-0002-8531-7386]
\ead{yaolei710@foxmail.com}
\fnmark[1]

\author[1]{Yikai Yang}
\fnmark[1]

\author[1]{Wenbo Peng}

\author[3,1]{Shumei Miao}

\author[1]{Yutao Hu}%
\cormark[1]

\author[1,2]{Guanyu Yang}[orcid=0000-0003-3704-1722]%
\ead{yang.list@seu.edu.cn}  
\cormark[1]







\affiliation[1]{
            organization={Key Laboratory of New Generation Artificial Intelligence Technology and Its Interdisciplinary Applications (Southeast University), Ministry of Education},
            addressline={No.2, Sipai Lou, Xuanwu District}, 
            city={Nanjing},
            postcode={210096}, 
            country={China}}

\affiliation[2]{
            organization={Jiangsu Province Joint International Research Laboratory of Medical Information Processing},
            city={Nanjing},
            postcode={210096}, 
            country={China}}

\affiliation[3]{
            organization={The First Affiliated Hospital of Nanjing Medical University},
            city={Nanjing},
            postcode={210029}, 
            country={China}}








\cortext[1]{Corresponding author}

\fntext[1]{Yaolei Qi and Yikai Yang contributed equally to this work.}


\begin{abstract}
Complex tubular structures are essential in medical imaging and computer-assisted diagnosis, where their integrity enhances anatomical visualization and lesion detection. 
However, existing segmentation algorithms struggle with structural discontinuities, particularly in severe clinical cases such as coronary artery stenosis and vessel occlusions, which leads to undesired discontinuity and compromising downstream diagnostic accuracy. Therefore, it is imperative to reconnect discontinuous structures to ensure their completeness. In this study, we explore the tubular structure completion based on point cloud for the first time and establish a Point Cloud-based Coronary Artery Completion (PC-CAC) dataset, which is derived from real clinical data. This dataset provides a novel benchmark for tubular structure completion. Additionally, we propose TSRNet, a Tubular Structure Reconnection Network that integrates a detail-preservated feature extractor, a multiple dense refinement strategy, and a global-to-local loss function to ensure accurate reconnection while maintaining structural integrity. Comprehensive experiments on our PC-CAC and two additional public datasets (PC-ImageCAS and PC-PTR) demonstrate that our method consistently outperforms state-of-the-art approaches across multiple evaluation metrics, setting a new benchmark for point cloud-based tubular structure reconstruction. Our benchmark is available at \url{https://github.com/YaoleiQi/PCCAC}.
\end{abstract}


\maketitle











\section{Introduction}
\label{sec:intro}
\begin{figure}[t!]
  \centering
   \includegraphics[width=0.8\linewidth]{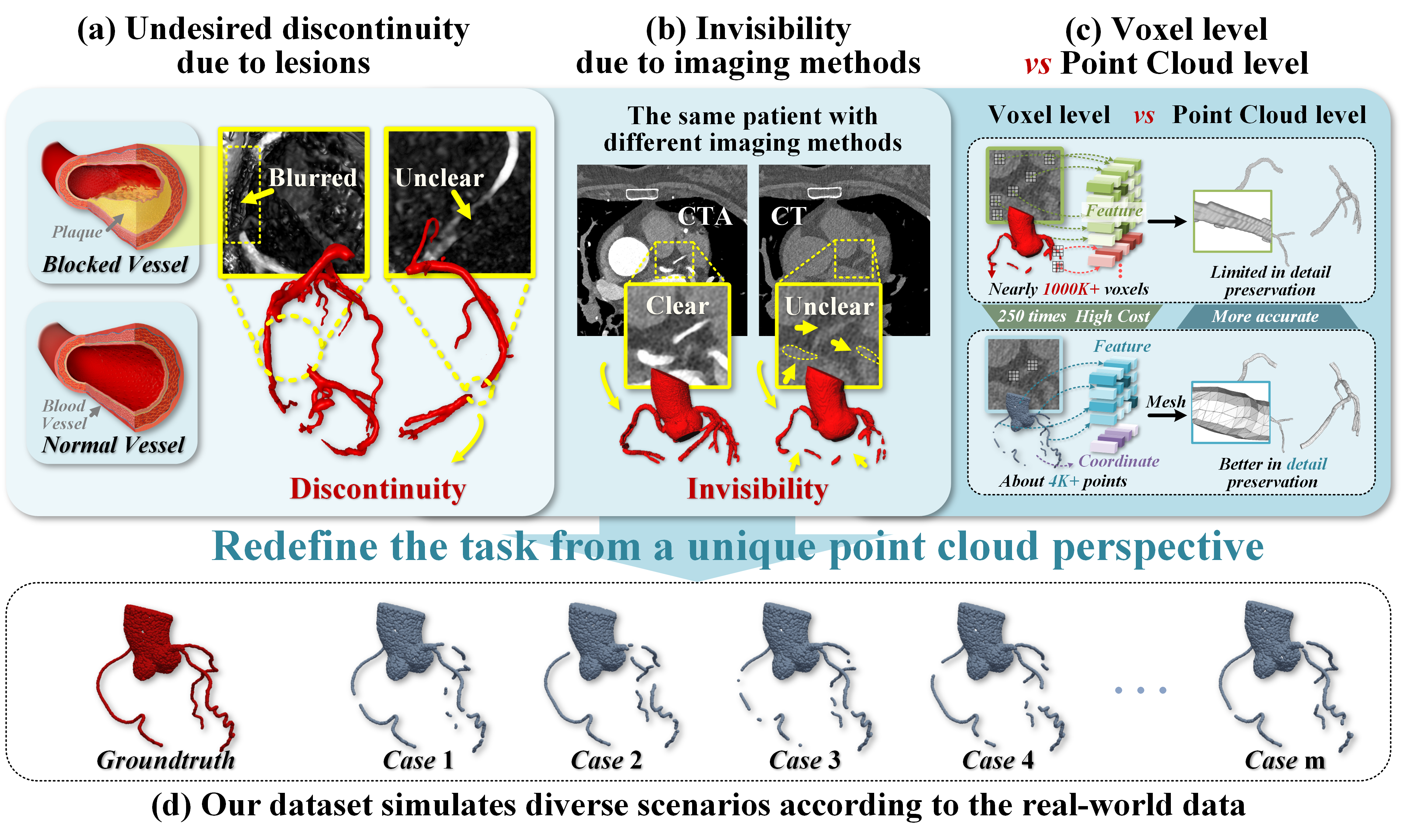}
   \caption{
   \textbf{Motivation.} (a) shows the undesired discontinuity caused by unclear image representation in structures affected by lesions. (b) shows vessels visualized across different modalities, where thin tubular structures are prone to being obscured, making them difficult to observe without the use of contrast agents. (c) shows the differences in structural reconnection from the perspectives of both voxel-based and point cloud representation. (d) our work redefines the reconnection task from a point cloud perspective and builds a dataset that is simulated from real clinical data.}
   \label{fig: Intro_one}
\end{figure}

\begin{figure}[t!]
  \centering
   \includegraphics[width=0.8\linewidth]{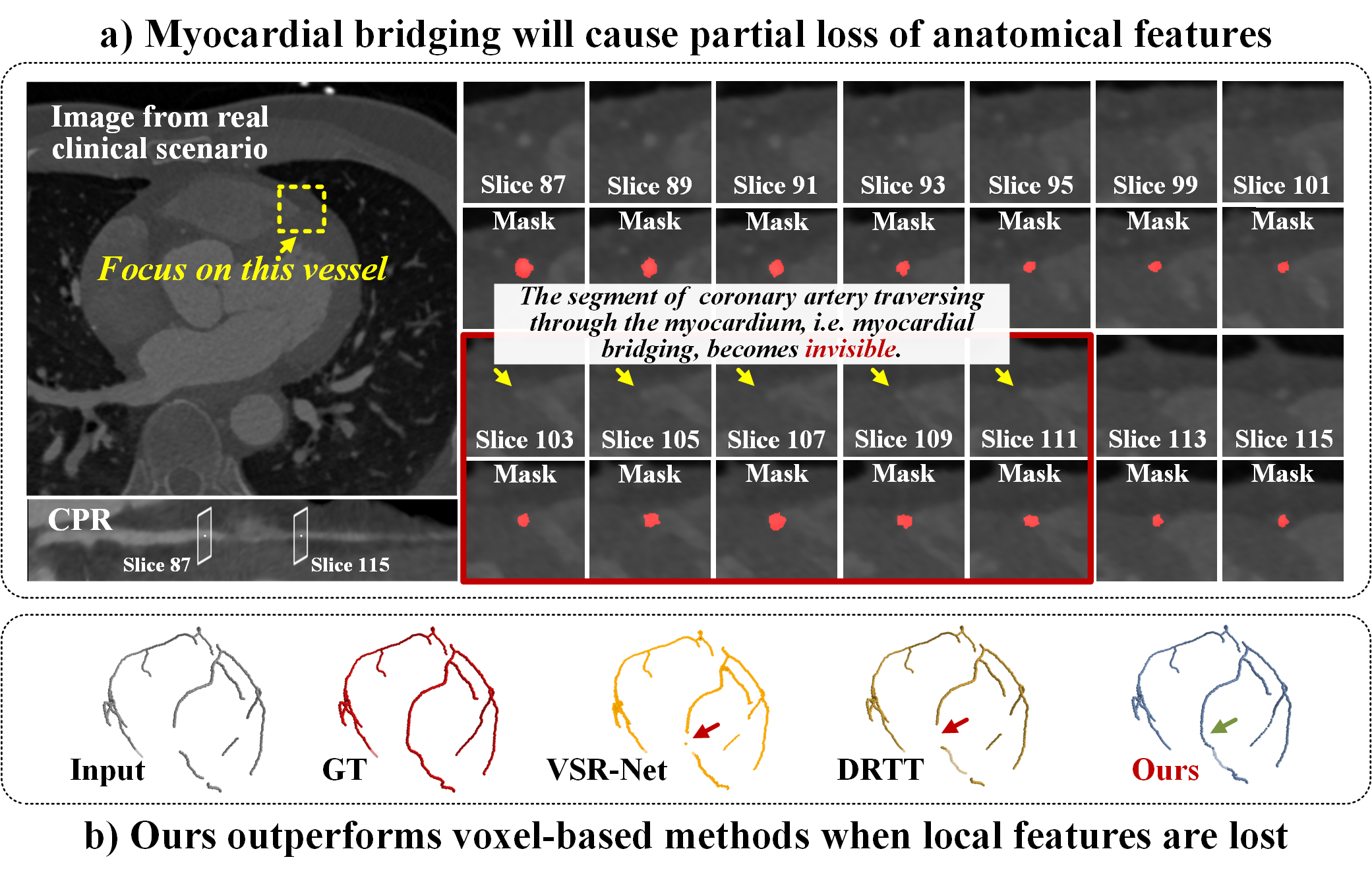}
   \caption{
   (a) Illustration of myocardial bridging in a real clinical case, where the lesion leads to local structural loss that challenges voxel-based models. (b) Our proposed method shows superior reconnection performance, and better handles such fracture compared to existing voxel-based approaches.}
   \label{fig: Intro_two}
\end{figure}

Complex tubular structures, such as blood vessels, constitute essential components of human physiology and play a critical role in medical imaging and computer-assisted diagnosis~\citep{li2022human}. Preserving their structural integrity is essential, as it enables the accurate visualization for radiologists and facilitates the detection of lesions~\citep{ko2017noninvasive}, supporting more informed and precise treatment decisions~\citep{gharleghi2022towards}. Existing segmentation algorithms~\citep{qi2023dynamic, kong2020learning, zhang2023anatomy, zhang2022progressive} have made progress in analyzing tubular structures. However, in more clinically significant and severe cases (Fig.~\ref{fig: Intro_one} (a)), such as coronary artery stenosis~\citep{madhavan2014coronary, pijls2012functional}, blockage~\citep{garje2015design}, and myocardial bridging (Fig.~\ref{fig: Intro_two} (a))~\citep{lee2015myocardial}, the continuity of segmentation is often compromised, resulting in fractures that negatively affect downstream diagnosis. Therefore, it is imperative to design an effective and robust reconstruction framework to preserve the anatomical integrity of complex tubular structures.

Recently, some studies~\citep{mou2019dense, qiu2023corsegrec, han2022reconnection, weng2023topology} have investigated the reconnection of fractured tubular structures using voxel-based approaches. However, these methods encounter notable limitations in handling the following complex scenarios. As shown in Fig.~\ref{fig: Intro_one} (a) and Fig.~\ref{fig: Intro_two} (a), lesions such as coronary artery stenosis cause blurry and unclear image representations, reducing the visibility of critical vascular structures. In cases of complete occlusion~\citep{thanvi2007complete}, this issue may further lead to missing information in local regions, resulting in undesired discontinuity. Such disruptions compromise the structural integrity of tubular anatomy and pose significant challenges for accurate reconnection. Similarly, as depicted in Fig.~\ref{fig: Intro_one} (b), variations across imaging modalities also present additional challenges. For patients allergic to contrast agents, non-contrast imaging presents substantial obstacles in analyzing tubular structures due to reduced visibility and lack of enhancement. These fine vascular features are often difficult to distinguish, which hampers visual assessment and complicates structural reconnection when relying solely on voxel-based methods.

Fortunately, point clouds, as a structured data representation that emphasizes coherence and consistency~\citep{fei2022comprehensive, chen2021shape, bernard2017shape,  beetz2023multi}, provide advantages that are absent at the voxel level. As shown in Fig.~\ref{fig: Intro_one} (c), point cloud-based methods offer notable advantages in computational efficiency and accuracy. Voxel-based methods process over 1,000K voxels, resulting in high computational cost, whereas point cloud-based methods operate on around 4K points. Additionally, the reconnected point cloud can then be directly utilized to construct a high-resolution sub-voxel mesh model, enabling accurate anatomical reconstruction and simulation without the need to revert to the original voxel representation. For instance, in coronary arteries, this enables the creation of a more accurate physical model for downstream tasks such as hemodynamic simulation~\citep{ko2017noninvasive}. Thus, we redefine the reconnection of complex tubular structures from a point cloud perspective (Fig.~\ref{fig: Intro_one} (d)) for the first time.

However, the direct application of point cloud completion to fractured tubular structures presents several key challenges:
(a) \textbf{Imbalanced point distribution}. The number and density of points vary greatly between large anatomical regions (e.g., the aorta) and slender elongated segments (e.g., coronary vessels). This extreme disparity biases the model toward larger structures and hinders its ability to capture fine structural details, ultimately degrading performance on thin tubular structures.
(b) \textbf{Complex global topology}. Tubular structures exhibit intricate and delicate morphologies across individuals. Compared to objects with regular or simple structures, predicting missing segments in thin tubular structures is particularly difficult, especially when capturing intricate curvature details. As a result, the model tends to produce incomplete or topologically incorrect structures during point generation.

To tackle the above obstacles, as a first attempt, we propose a novel \textbf{P}oint \textbf{C}loud-based \textbf{C}oronary \textbf{A}rtery \textbf{C}ompletion (PC-CAC) dataset and propose a robust, detail-preserved baseline for the precise reconnection of fractured tubular structures. 
(1) To enhance the integrity of complex tubular structures, we redefine the reconnection task from a point cloud perspective. To our best knowledge, we are the first to construct a dedicated dataset PC-CAC. Our dataset is derived from clinical data, incorporating segmentation results while preserving fractures, under-segmentation, and other imperfections. Furthermore, to simulate more challenging real-world scenarios, 10\%–30\% of the points from damage-prone regions within the main structures are removed from the input to better reflect the complexity of structural discontinuities. This design facilitates a more rigorous evaluation of completion algorithms and promotes the advancement of effective reconnection strategies.
(2) To address the challenges of imbalanced point distribution and complex global topology, we propose a \textbf{T}ubular \textbf{S}tructure \textbf{R}econnection Network (\ourmodel{}) designed for the accurate reconnection of fractured complex tubular structures, which comprises a detail-preserved feature extractor, a multiple dense refinement strategy, and a global-to-local loss function. These components cooperate to enhance detail preservation and effectively handle hard-to-represent regions.
(3) To objectively evaluate our approach, experiments are conducted on our PC-CAC dataset and two public datasets. We select a coronary artery segmentation (ImageCAS) dataset~\citep{zeng2023imagecas} and a pulmonary tree repairing (PTR) dataset~\citep{weng2023topology}. Then, we convert these two open-source datasets into point cloud format, namely PC-ImageCAS and PC-PTR, and conduct validation accordingly. 

In summary, our contributions are as follows:
\begin{itemize}
    \item \textbf{The first point cloud-based tubular structure reconnection dataset}: To our best knowledge, we construct the first point cloud-based coronary artery completion (PC-CAC) dataset from clinical data. Our benchmark provides a new perspective for tubular structure reconnection and fostering advancements in this field.
    \item \textbf{A novel exploration and high-performing baseline}: Our work represents the first attempt to explore tubular structure reconnection based on point cloud. We propose a baseline designed for accurately reconnecting fractured tubular structures, comprising a detail-preserved feature extractor, a multiple dense refinement strategy, and a global-to-local loss function. These methods cooperate to enhance detail preservation and effectively handle hard-to-represent regions.
    \item \textbf{A comprehensive evaluation across multiple datasets}. To objectively evaluate our approach, experiments are conducted on our PC-CAC dataset and two public datasets. Experimental results show that our method achieves state-of-the-art performance across multiple datasets.
\end{itemize}

\section{Related Works}
\label{sec:relatedworks}
\subsection{Reconnection methods from voxel perspective}
Many methods have been proposed to reconnect fractured vessels from the image perspective. (1) Traditional walk-based methods.~\citep{mou2019dense, qiu2023corsegrec, gupta2023topology}. Random walk classifies unlabeled pixels based on pre-annotated vessel (foreground) and background seed points, initially used directly for vessel segmentation~\citep{grady2006random, li2015automated}. Building on this,~\citep{mou2019dense} proposed the Probability Regularized Walk (PRW) algorithm for fractured vessel connection: labeling all connected regions via morphological operations, matching minimal-distance point pairs, and integrating local vessel direction modeling with segmentation network probability outputs to form "walking" connections.~\citep{qiu2023corsegrec} extended the PRW algorithm to 3D and applied it to coronary arteries.~\citep{gupta2023topology} introduced a perturb-and-map mechanism to diversify path search and avoid local optima or deviations. These methods rely on local information while neglecting global morphology and are limited by weak voxel-level features, heavily depending on endpoint detection accuracy. As a result, they struggle to maintain structural continuity, often leading to misaligned connections, incomplete reconstructions, or even false reconnection in complex structures. These limitations hinder their applicability in real-world clinical scenarios, where missing or noisy data further exacerbate the challenges. However, our approach integrates global and local information through point cloud level completion without requiring explicit fracture endpoint identification.

\subsection{Generic point cloud completion methods}
PointNet~\citep{qi2017pointnet} and PointNet++~\citep{qi2017pointnet++} advanced point cloud completion. Numerous deep learning-based methods~\citep{xie2020grnet, yu2021pointr, zhou2022seedformer, xiang2022snowflake, wen2022pmp, chen2023anchorformer, rong2024cra, wang2024pointattn} have emerged, typically using encoder-decoder architectures to map partial inputs to complete shapes.~\citep{xie2020grnet} regularized unordered point clouds via 3D grids as intermediate representations, leveraging structural and contextual neighbour information.~\citep{yu2021pointr} adopted a Transformer encoder-decoder architecture, converting point clouds into point proxies and designing geometry-aware modules to exploit 3D geometric inductive biases.~\citep{zhou2022seedformer} introduced Patch Seeds to capture global structures and local patterns.~\citep{wen2022pmp} predicted unique point displacement paths under distance constraints for strict correspondence learning.~\citep{wang2024pointattn} proposed geometric detail perception and self-feature augmentation units to establish structural relationships via attention mechanisms. These methods prioritize large surface optimization over slender tubular structures, thus failing to capture the intricate geometric details. The absence of dedicated mechanisms for preserving thin tubular features tends to result in disrupted topology, or over-reconstructions, ultimately compromising diagnostic reliability and downstream clinical applications. However, our approach specifically targets elongated vascular features through clinical data-driven training and continuity constraints for vascular topology.

\section{Methodology}
\label{sec:methodology}
\begin{figure*}[!t]
    \centering
    \includegraphics[width=\linewidth]{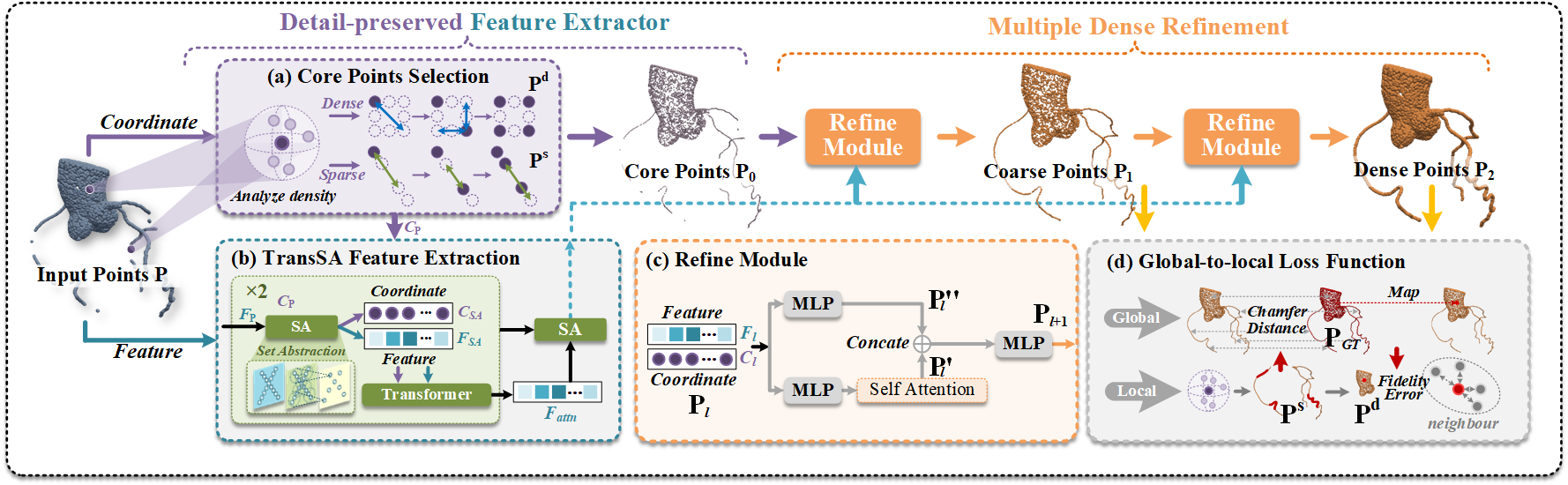}
    \caption{\textbf{Framework}. Our framework comprises two main stages: a Detail-preserved Feature Extractor and a Multiple Dense Refinement strategy. These are jointly optimized using a Global-to-local Loss Function to ensure the structural consistency. (a) Core Points Selection analyzes the density to extract critical points that are easily overlooked. (b) TransSA Feature Extraction utilizes a Transformer-based set abstraction to capture both local and global features. (c) Refine Module progressively refines results, generating coarse-to-fine reconstructions. (d) Global-to-local Loss constrains the reconstruction by enforcing global consistency via Chamfer Distance and enhancing local structure with Fidelity Error.}
    \label{fig3: method}
\end{figure*}

As shown in Fig.~\ref{fig3: method}, our \ourmodel{} comprises two main stages: a Detail-preserved Feature Extractor (Sec.~\ref{method:sec1}) and a Multiple Dense Refinement strategy (Sec.~\ref{method:sec2}). These are jointly optimized using a Global-to-local Loss Function (Sec.~\ref{method:sec3}) to ensure the structural consistency. To leverage geometric sparsity, the entire framework operates directly on point cloud inputs. The procedure for converting voxel-based segmentation results with anatomical discontinuities into point clouds is detailed in the next section (Sec.~\ref{sec:ExpSet}) and included in our open-source implementation.

\subsection{Detail-preserved Feature Extractor}
\label{method:sec1}
In this section, we discuss how to perform the Detail-preserved Feature Extractor. Given the input point cloud as $\mathbf{P} = \{\mathbf{p}_i \vert i = 1, 2, \ldots, N \} \in \mathbb{R}^{N\times3}$, where $N$ represents the total number of points, and each \begin{math}\mathbf{p}_i\end{math} contains 3D coordinates \begin{math}(\mathnormal{x}, \mathnormal{y}, \mathnormal{z})\end{math}. Within our network architecture, the extractor captures both global shape and local geometric context of the input point cloud $\mathbf{P}$ and downsamples it to support the subsequent coarse-to-fine refinement process. 

\subsubsection{Core Points Selection Module}
Progressive generation aims to achieve global densification of sparse point clouds. Thus, we first derive a sparse core point cloud $\mathbf{P}_0$ from the original $\mathbf{P}$ to integrate the reconnection of fracture locations into the global densification process. To address the challenge of imbalanced point distribution, which arises from the small spatial proportion occupied by complex tubular structures in the overall point cloud, we propose the Core Points Selection (CPS) module. CPS dynamically adjusts the attention of the model to different point types during training (Fig.~\ref{fig3: method} (a)). By analyzing density variations and employing adaptive selection strategies, CPS effectively identifies critical points, ensuring a well-balanced point distribution and mitigating the impact of sparsity.

Specifically, CPS first analyzes the density distribution of the input point cloud $\mathbf{P}$ to separate dense $\mathbf{P^d}$ and sparse regions $\mathbf{P^s}$. Then, the Farthest Point Selection~\citep{qi2017pointnet} is employed as the fundamental approach for core point selection, while we introduce adaptive constraints tailored to different density conditions to optimize the selection process in both dense and sparse regions. To ensure balanced and effective critical points selection, CPS incorporates three main constraints: (1) \textbf{Simultaneous selection in separate dense and sparse regions.} CPS operates separately on both dense and sparse regions, ensuring that the selection process adapts to local point distribution features without biasing towards high-density areas. (2) \textbf{Region-restricted point selection.} Points are selected strictly within their respective regions, meaning that points in high-density regions cannot be chosen from points in low-density regions. This constraint prevents structural distortion caused by disproportionate sampling across density variations. (3) \textbf{Priority on end and isolated points.} CPS prioritizes the retention of endpoints and isolated points, which are often crucial for maintaining continuity. These points play a key role in guiding the refinement process and preventing structural loss in later stages.

Under the condition of adhering to the principles, CPS starts from a single point and iteratively selects the most suitable point in different situations from the current point until the designated number $N_0$ of points is reached by $\mathbf{P_0} = \{\mathbf{P^d}, \mathbf{P^s}\} = C\!P\!S(\mathbf{P}, N_0)$, where $\mathbf{P^d}$ refers to the dense regions, and $\mathbf{P^s}$ refers to the sparse regions. $N_0$ is set to one quarter of the complete point cloud to facilitate high-quality refinement and to retain as much information from $\mathbf{P}$ as possible. $\mathbf{P}_0$ selected by CPS represent core points containing crucial information, exhibiting a uniform overall distribution, and preserving attention on the complex vascular structures.

\subsubsection{TransSA Feature Extraction Module.}
Inspired by~\citep{zhou2022seedformer}, which applies patch seeds to obtain rich information, we propose the TransSA module, which integrates Set Abstraction (SA) and Transformer~\citep{zhao2021point} to effectively extract features from irregular structures. 

As shown in Fig.~\ref{fig3: method} (b) and Eq.~(\ref{TransSA}), the SA block concatenates the input point coordinates $\mathbf{\mathit{C}_{P}}$ with their corresponding features $\mathbf{\mathit{F}_{P}}$, selects the centroid $\mathit{C}_{SA}$ within the local region, and identifies neighboring points of $\mathit{C}_{SA}$ via K-Nearest Neighbors (KNN)~\citep{zhang2017learning} to construct local representation. Subsequently, PointNet~\citep{qi2017pointnet} is applied to extract refined feature points $\mathit{F}_{SA}$. The Transformer module further processes the extracted features $\mathit{F}_{SA}$ and coordinates $\mathit{C}_{SA}$, mitigating the unordered nature of point cloud data while leveraging self-attention mechanisms for enhanced feature extraction, denoted as $\mathit{F}_{attn}$. Additionally, the SA module utilizes $\mathit{F}_{attn}$ to capture higher-dimensional representations, enriching the feature extraction process. By iteratively integrating Transformer with SA, the TransSA module effectively enhances feature discrimination, ensuring robust representation learning for complex structures.

\begin{equation}
\label{TransSA}
\begin{aligned}
    &\mathit{F}_{SA}, \mathit{C}_{SA} = SA(\mathbf{\mathit{F}_{P}}, \mathbf{\mathit{C}_{P}}), \\
    &\mathit{F}_{attn} = Transformer(\mathit{F}_{SA}, \mathit{C}_{SA}).
\end{aligned}
\end{equation}


\subsection{Multiple Dense Refinement Strategy}
\label{method:sec2}
As shown in Fig.~\ref{fig3: method}, the progressive network is designed to faithfully restore details from the sparse point cloud, gradually guiding the model to focus on details of completion. Multiple refine modules (Fig.~\ref{fig3: method} (c)) are designed to sequentially reconstruct the core point cloud $\mathbf{P}_0$ into the coarse point cloud $\mathbf{P}_1$, and then further refine $\mathbf{P}_1$ into the dense point cloud $\mathbf{P}_2$, which constitutes the final output. Features $\mathit{F}_\ell$ extracted by the TransSA module are incorporated into each refine module to better preserve the fidelity of the data. Within each refine module, the coordinates $\mathit{C}_\ell$ is fused with the feature $\mathit{F}_\ell$, which undergoes processing through an MLP followed by three rounds of Self-Attention, generating an intermediate representation $\mathbf{P}'_\ell$. Simultaneously, $\mathbf{P}_\ell$ is processed through another MLP to produce $\mathbf{P}''_\ell$. The final refined output $\mathbf{P}_{\ell+1}$ is obtained by applying an MLP to the concatenation of $\mathbf{P}'_\ell$ and $\mathbf{P}''_\ell$. The first refine module focuses on initial completion by addressing major discontinuities, while the second refine module further enhances the reconstructed details, emphasizing minute vascular structures to ensure finer structural integrity.

\subsection{Global-to-Local Loss Function}
\label{method:sec3}
Considering the progress of the coarse-to-fine training process, it becomes imperative to impose stage-specific constraints, gradually guiding the model optimization towards the desired direction. Therefore, the overall loss function $\mathcal{L}$ encompasses two components, global $\mathcal{L}_{g}$ and local $\mathcal{L}_{l}$, which are tailored to the differences in the outputs across different stages. In detail, $\ell_1$ Chamfer Distances ($C\!D^{\ell_1}$)~\citep{fan2017point} is employed as the primary metric, which utilizes $\ell_1$-norm to compute the distance between two sets of points, ensuring the precise global structural alignment. Furthermore, to maintain a high degree of similarity between the reconstructed output and the original input, and enhance the focus of the model on local regions, we integrate Fidelity Error ($F\!E$)~\citep{yuan2018pcn} into the loss formulation. Since $F\!E$ is defined as the average distance from each point in the input to its nearest neighbor in the output, effectively measuring how well the original structure is preserved during completion. 

In our study, loss calculation is divided into global and local perspectives. Specifically, the global perspective (Eq.~(\ref{eq: Lg})) focuses on the overall structural restoration, ensuring the coherence of the reconstructed shape. Meanwhile, the local perspective (Eq.~(\ref{eq: Ll})) leverages the CPS module mentioned above to analyze point set density and processes different density regions separately. This approach dynamically balances the representation of sparse and dense point sets, enhancing the model’s attention to sparse structures. The loss from global $\mathcal{L}_g$ becomes:

\begin{equation}
\label{eq: Lg}
    \mathcal{L}_g =                		
    \underbrace{C\!D^{\ell_1}(\mathbf{P}_1,\mathbf{P}_{G\!T})+F\!E(\mathbf{P}_1,\mathbf{P}_{G\!T})}_{\mathcal{L}_{coarse}}+           		\underbrace{C\!D^{\ell_1}(\mathbf{P}_2,\mathbf{P}_{G\!T})}_{\mathcal{L}_{fine}},
\end{equation}

\noindent
where $C\!D^{\ell_1}$ represents the $\ell_1$ Chamfer Distance, which measures the discrepancy between two point sets. $\mathcal{L}_g$ denotes the loss function from global perspective, consisting of $\mathcal{L}_{coarse}$ and $\mathcal{L}_{fine}$, corresponding to different stages. $\mathbf{P}_1$ and $\mathbf{P}_2$ represent the reconstructed point clouds at the coarse and fine stages, respectively, while $\mathbf{P}_{G\!T}$ denotes the ground truth point cloud. $F\!E$ refers to Fidelity Error, which evaluates how well the reconstructed structure preserves the original input, focusing on fine-grained details.

From the local perspective, the CPS module mentioned above is utilized to analyze the density distribution of the input point set $\mathbf{P}$, and separate it into two regions, including $\mathbf{P}^d$ for high-density areas and $\mathbf{P}^s$ for low-density areas, express as: $\{\mathbf{P}^d, \mathbf{P}^s\} = C\!P\!S(\mathbf{P}, N_0)$. Then $C\!P\!S(\cdot)$ is performed simultaneously on the output point sets $\mathbf{P}_1$, $\mathbf{P}_2$, and the ground truth $\mathbf{P}_{G\!T}$. Then the following loss calculation exclusively on $\mathbf{P}^s$ region:

\begin{equation}
\label{eq: Ll}
    \mathcal{L}_{l} =		
    \underbrace{C\!D^{\ell_1}(\mathbf{P}_1^s,\mathbf{P}_{G\!T}^s) + F\!E(\mathbf{P}_1^s,\mathbf{P}_{G\!T}^s)}_{\mathcal{L}_{coarse}} + 
    \underbrace{C\!D^{\ell_1}(\mathbf{P}_2^s,\mathbf{P}_{G\!T}^s)}_{\mathcal{L}_{fine}},
\end{equation}

\noindent
where $C\!D^{\ell_1}$ and $F\!E$ are consistent with Eq.~(\ref{eq: Lg}). Specifically, $\mathbf{P}_1^s$, $\mathbf{P}_2^s$ and $\mathbf{P}_{G\!T}^s$ represent the sparse regions within the point sets $\mathbf{P}_1$, $\mathbf{P}_2$ and $\mathbf{P}_{G\!T}$, respectively.

Finally, the global-to-local training process is regulated by a threshold $\varepsilon$, ensuring a stage-aware transition between loss components. The complete formulation of the loss function $\mathcal{L}$ is defined as follows:

\begin{equation}
\mathcal{L}= 
\begin{cases}
\mathcal{L}_{g}, & \mathcal{L}_{g} > \varepsilon,
\\
(1-\gamma)\mspace{3mu} \mathcal{L}_{g} + \gamma\mspace{3mu} \mathcal{L}_{l}, &\text{otherwise},
\end{cases}
\end{equation}

\noindent
where $\mathcal{L}_{g}$ represents the global loss, ensuring overall structural consistency, while $\mathcal{L}_{l}$  accounts for finer details by focusing on local reconstruction accuracy. The threshold $\varepsilon$ acts as a phase transition parameter, determining whether the loss function should prioritize global structural alignment or incorporate local refinements. The trade-off parameter $\gamma$ dynamically adjusts the weighting between global and local losses when the global loss $\mathcal{L}_{g}$ is below the threshold, allowing a gradual shift in emphasis as training progresses. This adaptive loss strategy ensures that early training stages prioritize overall geometric consistency, while later stages refine fine details, leading to progressive, high-fidelity reconstructions of complex tubular structures.

\section{Experimental Setting}
\label{sec:ExpSet}
\subsection{Datasets Description}
As shown in Tab.~\ref{tab: data}, to objectively evaluate our approach, experiments are conducted on our PC-CAC dataset and two public datasets. We select a coronary artery segmentation (ImageCAS) dataset~\citep{zeng2023imagecas} and a pulmonary tree repairing (PTR) dataset~\citep{weng2023topology}. Then, we convert these two open-source datasets into point cloud format, namely PC-ImageCAS and PC-PTR, and conduct validation accordingly.

\textbf{PC-CAC Dataset}. We construct the \textbf{P}oint \textbf{C}loud-based \textbf{C}oronary \textbf{A}rtery \textbf{C}ompletion (PC-CAC) dataset, derived from real clinical coronary artery data, comprising a total of 427 patients (3,416 cases). These images were scanned on a SOMATOM Definition Flash and the contrast media was injected during the scanning process. The x/y-resolution of these CT images is between 0.25 $\sim$ 0.57 $mm$/voxel and the slice thickness is between 0.75 to 3 $mm$/voxel. The x/y-size of the images is 512 voxels and the z-size is between 128 $\sim$ 994 voxels. For pre-processing, we first resample the resolution of these images to $1mm \times 1mm \times 1mm$ for a unified resolution, then threshold their grayscale value to [0, 2048] and normalize them to [0, 1] for unified intensity. 

The dataset is partitioned into 300 patients (2,400 cases) for training, 40 patients (320 cases) for validation, and 87 patients (696 cases) for testing. Each point cloud consists of 4,096 points, representing the aorta and the left and right coronary arteries, where the aorta contains 3,072 points, and the coronary arteries contain 1,024 points. The input point clouds in our dataset originate from real segmentation results~\citep{isensee2021nnu, qi2023dynamic} using Alg.~\ref{alg1}. To further simulate real-world challenges, we introduce synthetic fractures at high-risk locations, including bifurcations and tortuous structures, where discontinuities are most likely to occur. Additionally, we incorporate random fractures and noise to enhance robustness and generalization. To ensure a diverse and challenging reconstruction task, we generate 8 distinct input cases per patient, capturing a wide range of vascular conditions. To simulate fractures, 10\%–30\% of the points in the coronary arteries are removed, while the aortic structure remains unchanged, ensuring a realistic and challenging reconstruction scenario. 


\textbf{PC-ImageCAS Dataset}. To further enhance the evaluation of our proposed approach, we introduce a public coronary artery segmentation dataset ImageCAS~\citep{zeng2023imagecas}, and convert it into a point cloud format called PC-ImageCAS dataset. The dataset consists of 8,000 cases, derived from 1,000 patients, where each complete point cloud sample captures detailed anatomical structures. The dataset is partitioned into 700 patients (5,600 cases) for training, 100 patients (800 cases) for validation, and 200 patients (1,600 cases) for testing, maintaining consistency in dataset division. This structured partitioning ensures a comprehensive assessment across varying clinical conditions, reinforcing the reliability of our framework in real-world applications.

\textbf{PC-PTR Dataset}. To further assess the generalization capability of our proposed baseline, we utilize the pulmonary arteries in Pulmonary Tree Repairing (PTR) dataset~\citep{weng2023topology} and convert them into point clouds format, referred to as PC-PTR. The dataset comprises 799 patients (6,392 cases), with each sample consisting of 8,192 points, where 5\%–15\% of the points are deliberately removed to simulate structural degradation. The dataset is partitioned into 559 patients (4,472 cases) for training, 80 patients (640 cases) for validation, and 160 patients (1,280 cases)  for testing, following the same as the original PTR dataset.

\begin{table}[!t]
\caption{The details of our dataset from the real clinical scenario and two public available datasets in our verification tasks.} 
\begin{center}
\resizebox{\textwidth}{!}{
\begin{threeparttable} 
\begin{tabular}{|c|c|c|c|c|} 
\hline
\multicolumn{5}{|c|}{a) The details of two public available datasets in our task} \\

\hline
\textbf{Name} & \textbf{Target dataset} & \textbf{Train/Val/Test} 
& \textbf{Detail information} & \textbf{Pre-processing} \\

\hline
PC-ImageCAS & ImageCAS~\citep{zeng2023imagecas} $^a$
& \makecell[c]{
700/100/200 \\
5600/800/1600 $^1$ \\
}
& \makecell[l]{
1. Scanner: Siemens 128-slice dual-source \\
2. Planar resolution: 0.29 $\sim$ 0.43 $mm^2$ \\
3. Slice thickness: 0.25 $\sim$ 0.45 $mm$ \\
4. x/y-size: 512 voxels, z-size: 206 $\sim$ 275 voxels \\
             }   
& \makecell[l]{
1. Resample the resolution to 1 $mm^3$\\
2. Normalize via $\frac{max(min(0, x), 2048)}{2048}$ \\
3. Obtain segmentation results \\
4. Extract \textbf{\textit{aorta}} and main \textbf{\textit{coronary}} branches \\
5. Generate point cloud based on surface of \textbf{\textit{aorta}} \\    
6. Generate point cloud based on centerline of \textbf{\textit{coronary}} \\   
} \\

\hline
PC-PTR & PTR~\citep{weng2023topology} $^b$ 
& \makecell[c]{
599/80/160 \\
4472/640/1280 $^1$ \\
}
& \makecell[l]{
1. Scan from multiple medical centers \\
2. Resolution: already be processed to 1 $mm^3$\\
3. x/y-size: 512 voxels, z-size: 177 $\sim$ 798 voxels \\
}   
& \makecell[l]{  
Generate point cloud based on centerline of vessels \\   
} \\

\hline
\multicolumn{5}{|c|}{b) The details of our proposed dataset from real clinical data} \\

\hline
\textbf{Name} & \textbf{Target dataset} & \textbf{Train/Val/Test} 
& \textbf{Detail information} & \textbf{Pre-processing} \\

\hline
PC-CAC & PC-CAC
& \makecell[c]{
300/40/87 \\
2400/320/696 $^1$ \\
}
& \makecell[l]{
1. Scanner: SOMATOM Definition Flash \\
2. x/y-resolution: 0.25 $\sim$ 0.57 $mm$/voxel \\
3. Slice thickness: 0.75 $\sim$ 3 $mm$/voxel \\
4. x/y-size: 512 voxels, z-size: 128 $\sim$ 994 voxels \\
             }   
& \makecell[l]{
1. Resample the resolution to 1 $mm^3$\\
2. Normalize via $\frac{max(min(0, x), 2048)}{2048}$ \\
3. Obtain segmentation results \\
4. Extract \textbf{\textit{aorta}} and main \textbf{\textit{coronary}} branches \\
5. Generate point cloud based on surface of \textbf{\textit{aorta}} \\    
6. Generate point cloud based on centerline of \textbf{\textit{coronary}} \\   
} \\

\hline

\end{tabular}
\begin{tablenotes}
    \item $^1$ To ensure a diverse and challenging reconstruction task, each patient generates 8 distinct input cases, capturing a wide range of conditions.
    \item $^a$ ImageCAS: \url{https://github.com/XiaoweiXu/ImageCAS-A-Large-Scale-Dataset-and-Benchmark-for-Coronary-Artery-Segmentation-based-on-CT}.
    \item $^b$ PTR: \url{https://github.com/M3DV/pulmonary-tree-repairing}.
\end{tablenotes}
\end{threeparttable}
} 

\end{center}
\label{tab: data}
\end{table}

\subsection{Details of Voxel-to-Point Clouds Process}
This section discusses how to perform the voxel-to-point cloud conversion process. Using our proposed coronary artery dataset as an example, it consists of segmentation results (using segmentation models such as nnU-Net~\citep{isensee2021nnu, qi2023dynamic}) that preserve fractures and under-segmentation, along with data that simulate challenging real-world scenarios. Regarding the segmentation results, the skeleton is extracted from the voxels, followed by a point-by-point mapping to form a point cloud. For the simulation of fractures, the complete procedure is illustrated in Alg.~\ref{alg1}. The operations on the segmentation results are similar to those of this algorithm, except that simulated fractures are not required. The input point cloud simulation construction process involves four steps: obtaining segmentation or annotation, converting into point clouds, normalizing, and simulating fractures. To better simulate real-world scenarios, operations such as removing cavities from voxels and extracting the main trunk from point clouds are also performed.

\IncMargin{1em}
\begin{algorithm}[htbp]
\small
\SetKwData{Left}{left}\SetKwData{This}{this}\SetKwData{Up}{up} \SetKwFunction{Union}{Union}\SetKwFunction{FindCompress}{FindCompress} \SetKwInOut{Input}{input}\SetKwInOut{Output}{output}
	
    \Input{The annotation of voxel-based \textbf{\textit{aorta}} $y^a$ and \textbf{\textit{coronary}} $y^c$; \\
               Numbers of patients $K$, numbers of points $N$, number of cases for each simulation $M$.} 
    \Output{The ground truth of point cloud-based vessels (including \textbf{\textit{aorta}} and \textbf{\textit{coronary}}) $\mathbf{P}_{G\!T}$; \\
                The input of point cloud-based vessels (including \textbf{\textit{aorta}} and \textbf{\textit{coronary}}) $\mathbf{P}$. }
	 \BlankLine 

    \emph{Eliminate cavities in $y^c$}\;
     Define the spherical structuring element $e$ to control the extent of elimination;\\
    Apply morphological operations (Dilation, Erosion) with structuring element $e$;\\

    \emph{Obtain centerline point cloud $y''^c$ of \textbf{\textit{coronary}}}\;

    \emph{Obtain surface point cloud $y''^a$ of \textbf{\textit{aorta}}}\;
        \For{$k=1\dots K$}{ 
            $y'^a$ = MC($y^a$); \Comment{MC represents the Marching Cubes. $y'^a$ refers to the point cloud surface of \textbf{\textit{aorta}}.} \\
            $N_a$ = $N$ - PointNums($y''^c$) \\
            $y''^a$ = FPS($y'^a, N_a$) \Comment{FPS represents the Farthest Point Sampling.}
      }

    \emph{Obtain main trunk point cloud $y^t$ of \textbf{\textit{coronary}}}; \Comment{Remove irrelevant branches.}\\
    \For{$k=1\dots K$}{ 
        \emph{Calculate all pairs of endpoints ($t_1$, $t_2$) for $y''^c$}\;
        \For{each endpoint pair ($t_1$, $t_2$)}{
            Perform Depth-First Search from $t_1$ to $t_2$, updating the shortest path\;
            $i$ = GetIndex($t_1$, $t_2$); \Comment{Assign index $i$ for each pair.} \\
            Record the shortest path $y^t_i$.\\
        } 
        $y^t$ = Union\{$y^t_i\}$.
    }
    
    \emph{Normalize the point clouds $y^t$ and $y''^a$}\;
    \For{$k=1\dots K$}{ 
        Calculate the centroid of Union\{$y^t$, $y''^a$\}\;
        \For{each point $p \in$ Union\{$y^t$, $y''^a$\}}{
            Subtract the centroid from point $p$\;
             Scale the points to fit within the range $[-1, 1]$.\\
        }
    }

\emph{Obtain ground truth point cloud $\mathbf{P}_{G\!T}$ and simulated fractured input point clouds $\mathbf{P}$}\;
    
    \For{$k=1\dots K$}{ 
        $\mathbf{P}_{G\!T}$ = Union\{$y''^a$, $y^t$\}, $\mathbf{P}$ = $\emptyset$\; 
        \For{$m=1\dots M$}{  
            $R$ = Random(min\_ratio, max\_ratio); \Comment{Proportion of points to be removed.}\\
            $B$ = Random(min\_break, max\_break); \Comment{Number of regions to break.}\\
            Randomly partition $y^t \times R$ points into $B$ groups and store them in $\text{remove}[B]$\;
            $y^t_{\text{copy}}$ = Copy($y^t$)\;
            \For{$b=1\dots B$}{  
                $\text{connect}$ = GetConnectedComponents($y^t_{\text{copy}}$)\; 
                Randomly delete a connected component of size $\text{remove}[b]$ from $y^t_{\text{copy}}$\;
                $\text{connect}'$ = GetConnectedComponents($y^t_{\text{copy}}$)\; 
                \If{$\text{connect}' == \text{connect}$}{  
                    Retry the current deletion; \Comment{Did not cause a fracture, retry.} \\
                }
            }
            $\mathbf{P}$ = $\mathbf{P} \cup \{y^t_{\text{copy}}\}$; \\
        }
    }
    \emph{End, ground truth $\mathbf{P}_{G\!T}$ and $M$ sets of input point clouds $\mathbf{P}$ are obtained.}\\

    \caption{Conversion of Coronary Artery Voxels into Corresponding Point Cloud from Clinical Data.}
    \label{alg1} 
\end{algorithm}
\DecMargin{1em} 

\subsection{Experimental Configurations}
\textbf{Evaluation Settings}. Our evaluation is divided into two main parts: 1) comparisons with point cloud-based methods (Sec.~\ref{res:pointcloud}), and 2) comparisons with voxel-based methods (Sec.~\ref{res:voxel}). For both parts, we conduct comprehensive performance evaluations as well as ablation studies to validate the effectiveness of our proposed \ourmodel{}. For \textbf{\textit{point cloud-based methods}}, we select classic point cloud completion algorithms GRNet~\citep{xie2020grnet} and PointTR~\citep{yu2021pointr}, as well as the methods such as SeedFormer~\citep{zhou2022seedformer} and SnowflakeNet~\citep{xiang2022snowflake} proposed in 2022, AnchorFormer~\citep{chen2023anchorformer} proposed in 2023, PointAttN~\citep{wang2024pointattn} and CRA-PCN~\citep{rong2024cra} proposed in 2024. For \textbf{\textit{voxel-based methods}}, we adopt DRTT~\citep{li2021deep} and VSRNet~\citep{ye2025vsr} as representative methods, both of which have been used in volumetric vascular reconstruction tasks. All models are trained on the same datasets under identical implementation settings to ensure a fair comparison.

\textbf{Evaluation Metrics}. All metrics are calculated for each case and averaged. 1) $\ell_1$ Chamfer Distances ($C\!D^{\ell_1}$)~\citep{fan2017point}. It uses $\ell_1$-norm to calculate the sum of the average closest distance between two point clouds. 2) F1-Score (F1)~\citep{tatarchenko2019single}. It evaluates the distance between the surfaces of objects. 3) Fidelity Error~\citep{yuan2018pcn}. It is used to measure how well the input is preserved, which calculates the average distance between the point and the nearest neighbor.

\textbf{Implementation Details}. Experiments are implemented using PyTorch on GeForce RTX 2080 Ti. Adam optimizer is used to train the models with an initial learning rate of \begin{math}10^{-4}\end{math}. The threshold parameter $\varepsilon$ for the loss function $\mathcal{L}$ is set to \begin{math}6\times10^{-3}\end{math} and the trade-off parameter $\gamma$ is set to 0.5. To ensure a more objective evaluation and to ignore the effect of inflated metrics caused by the high number of points in the aorta, the \textbf{\textit{coronary arteries are assessed independently}}.

\section{Results and Analysis}
\subsection{Compared with point cloud-based methods}
\label{res:pointcloud}
In this section, we present a comprehensive comparison between our \ourmodel{} and the state-of-the-art point cloud-based completion methods. The evaluation is conducted on three benchmark datasets: PC-CAC, PC-ImageCAS, and PC-PTR. We report both quantitative and qualitative results, followed by detailed ablation studies to assess the contribution of each component in our framework. The analysis focuses on three key performance indicators.

\subsubsection{Quantitative Evaluation}
\textbf{Comparative Results.}
The advantages of our \ourmodel{} on each metric are demonstrated in Table 2, and the results show that our framework consistently outperforms existing approaches across all three datasets: PC-CAC, PC-ImageCAS, and PC-PTR, achieving state-of-the-art performance in $C\!D^{\ell_1}$, F1, and Fidelity Error. Specifically, our method attains the lowest $C\!D^{\ell_1}$, demonstrating its effectiveness in accurately reconstructing fine-grained structural details. Additionally, our model achieves the highest F1 on PC-CAC and PC-PTR, while maintaining the second-best F1 on PC-ImageCAS. Furthermore, our approach consistently outperforms other methods in Fidelity Error, with the lowest values observed across all datasets. Notably, the most striking improvement is in Fidelity Error in PC-PTR which consists of dense and numerous elongated tubular structures, where our model achieves 0.102, marking a \textbf{\textit{97.2\% reduction}} compared to the second-best model SeedFormer, demonstrating unparalleled structural preservation. This confirms that our model preserves the original structure more faithfully while reducing unwanted distortions. Compared to other recent methods such as AnchorFormer, PointAttN and CRA-PCN, our framework achieves more stable and better performance, particularly in complex vascular structures from PC-PTR, where detail preservation is critical. These results validate the effectiveness of our detail-preserved feature extractor, multiple dense refinement strategy, and global-to-local loss function, collectively contributing to the superior performance of our method.

\begin{table}
\begin{center}
  \caption{Quantitative results of different point cloud-based methods on datasets PC-CAC, PC-ImageCAS and PC-PTR. \textbf{Bold} indicates the best results, and \underline{Underlined} indicates the second-best results.
}  
\resizebox{\textwidth}{!}{
  \begin{tabular}{c|ccc|ccc|ccc}

\hline   
\multirow{2}{*}{Method} 
&\multicolumn{3}{|c|}{\textbf{PC-CAC}} 
&\multicolumn{3}{|c|}{\textbf{PC-ImageCAS}}
&\multicolumn{3}{|c}{\textbf{PC-PTR}}\\
\cline{2-10}
 &$CD^{\ell_1}(10^{-3})\downarrow$ &F1$(\%)\uparrow$ &Fidelity$(10^{-3})\downarrow$ 
 &$CD^{\ell_1}(10^{-3})\downarrow$ &F1$(\%)\uparrow$ &Fidelity$(10^{-3})\downarrow$ 
 &$CD^{\ell_1}(10^{-3})\downarrow$ &F1$(\%)\uparrow$ &Fidelity$(10^{-3})\downarrow$ \\
    
\hline  

GRNet~\cite{xie2020grnet}
& 32.683$\pm$9.488 & 32.60$\pm$4.95 & 52.549$\pm$17.918
& 28.953$\pm$5.411 & 46.12$\pm$4.15 & 46.292$\pm$10.401
& 11.641$\pm$0.836 & 58.81$\pm$4.57 & 13.900$\pm$0.856 \\

PoinTr~\cite{yu2021pointr} 
& 4.452$\pm$0.896 & 87.98$\pm$10.76 & 4.335$\pm$1.634
& 4.861$\pm$0.920 & 85.83$\pm$12.61 & 4.796$\pm$1.422
& 6.129$\pm$1.510 & 80.80$\pm$15.72 & 5.203$\pm$2.572 \\

SeedFormer~\cite{zhou2022seedformer} 
& \underline{4.139$\pm$1.286} & \underline{93.41$\pm$2.84} & \underline{3.598$\pm$1.368}
& \underline{2.799$\pm$0.712} & \textbf{97.47$\pm$1.88} & \underline{2.563$\pm$0.825}
& \underline{4.803$\pm$1.356} & \underline{88.82$\pm$7.02} & \underline{3.666$\pm$0.914} \\

SnowflakeNet~\cite{xiang2022snowflake} 
& 5.661$\pm$1.661 & 88.11$\pm$3.37 & 5.816$\pm$2.365
& 2.979$\pm$0.893 & 95.89$\pm$2.54 & 2.836$\pm$1.288
& 6.033$\pm$1.327 & 83.06$\pm$6.97 & 5.666$\pm$1.009 \\

AnchorFormer~\cite{chen2023anchorformer} 
& 6.747$\pm$0.871 & 83.11$\pm$7.32 & 6.178$\pm$0.845
& 4.699$\pm$0.435 & 87.68$\pm$11.62 & 4.392$\pm$0.693
& 11.491$\pm$1.150 & 48.48$\pm$7.51 & 9.237$\pm$1.107 \\

PointAttN~\cite{wang2024pointattn} 
& 8.288$\pm$1.348 & 67.22$\pm$4.47 & 6.660$\pm$2.488 
& 5.757$\pm$1.422 & 83.84$\pm$4.01 & 4.595$\pm$2.587
& 11.387$\pm$1.777 & 54.67$\pm$7.60 & 9.324$\pm$1.657 \\
   
CRA-PCN~\cite{rong2024cra}
& 5.869$\pm$1.100 & 88.75$\pm$3.87 & 6.245$\pm$2.171 
& 3.526$\pm$0.749 & 93.54$\pm$4.44 & 4.414$\pm$1.650
& 9.088$\pm$0.982 & 66.01$\pm$7.20 & 8.950$\pm$2.114 \\

\textbf{\ourmodel{} (Ours)} 
& \textbf{3.783$\pm$1.456} & \textbf{94.58$\pm$2.28} & \textbf{3.318$\pm$1.768} 

& \textbf{2.395$\pm$0.716} & \underline{96.83$\pm$1.94} & \textbf{2.062$\pm$0.910}

& \textbf{2.382$\pm$0.533} & \textbf{96.31$\pm$1.83} & \textbf{0.102$\pm$0.031} \\

\hline  
\end{tabular}
}
\end{center}

  \label{tab1}
\end{table}

\textbf{Ablation Study.}
The results in Table~\ref {tab2} demonstrate that each module contributes to the overall performance, with the best results achieved when all three components are combined (Row D). By comparing Row A (baseline) and Row B (with the Multiple Dense Refinement Strategy enabled), we observe a substantial improvement in $C\!D^{\ell_1}$ from 9.137 to 4.763, and an increase in F1 from 73.90\% to 93.27\%. This confirms that the Multiple Dense Refinement Strategy effectively enhances the completeness and precision of the reconstructed structures by progressively refining missing regions. Enabling the Global-to-Local Loss Function further improves performance, as seen in Row C, where $C\!D^{\ell_1}$ decreases to 3.877, F1 rises to 94.39\%, and Fidelity Error is also notably reduced, indicating that this loss function helps balance global structural consistency while preserving fine details. The integration of the Detail-Preserved Feature Extractor (Row D) leads to the best overall performance, achieving the lowest $C\!D^{\ell_1}$ of 3.783, lowest Fidelity Error of 3.318, and highest F1 of 94.58\%, which demonstrates that this module plays a crucial role in accurately reconstructing fine-grained tubular structures.



\begin{table}[!t]
    \centering
    \caption{Quantitative results of ablation analysis of different modules on our PC-CAC. \ding{172} represents the Multiple Dense Refinement Strategy. \ding{173} represents the Global-to-Local Loss Function. \ding{174} represents the Detail-Preserved Feature Extractor. }
    \resizebox{0.5\textwidth}{!}{
    \begin{tabular}{c|c|c|c|ccc}
        \hline
        & \ding{172} & \ding{173} & \ding{174} & $CD^{\ell_1}(10^{-3})\downarrow$ & F1(\%)$\uparrow$ & Fidelity$(10^{-3})\downarrow$ \\
        \hline
        A & & & & $9.137\pm7.993$ & $73.90\pm31.36$  & $7.950\pm7.577$ \\
        B & \checkmark & & & $4.763\pm0.593$ & $93.27\pm3.40$  & $4.809\pm0.654$\\
        C & \checkmark & \checkmark & & $3.877\pm1.684$ & $94.39\pm2.45$ & $3.351\pm1.809$ \\
        \rowcolor{gray!10}
        \textbf{D} & \checkmark & \checkmark & \checkmark & $\mathbf{3.783\pm1.456}$ & $\mathbf{94.58\pm2.28}$ & $\mathbf{3.318\pm1.768}$ \\
        \hline
    \end{tabular}
    }
    \label{tab2}
\end{table}

\textbf{Hyperparameter Ablation.}
To evaluate the impact of key hyperparameters in TSRNet, we perform a comprehensive ablation study on the PC-CAC dataset, as summarized in Table~\ref{tab3}. Specifically, we vary the global loss threshold $\epsilon$, the trade-off parameter $\gamma$, and the number of stacked SA+Transformer (S+T) blocks. Notably, Row 7 achieves the best overall performance with $\epsilon=2.7$  and $\gamma=0.5$.  We also explored the number of SA+Transformer blocks, the choice of $2$ achieves the best result, and is thus adopted in our model.

\begin{table}[htb]
\centering
\caption{Ablation study of key hyperparameters. We vary the global loss threshold $\epsilon$, the trade-off parameter $\gamma$, and the number of SA+Transformer (S+T) blocks. The best performance is highlighted in bold.}
\resizebox{0.5\textwidth}{!}{
\begin{tabular}{c|ccc|c|c|c}
\toprule
&$\epsilon$ & $\gamma$ & (S+T) &
$CD^{\ell_1}(10^{-3})\downarrow$ &
F1(\%)$\uparrow$ &
Fidelity($10^{-3}$)$\downarrow$ \\
\midrule
$1$&$3.1$   & $0.5$  & $2$ & $2.485 \pm 0.718$ & $96.06 \pm 2.11$ & $2.254 \pm 0.970$ \\
$2$&$2.9$   & $0.5$  & $2$  & $2.405 \pm 0.716$ & $96.60 \pm 1.97$ & $2.084 \pm 0.912$ \\
$3$&$2.7$   & $0.3$  & $2$  & $2.502 \pm 0.719$ & $96.56 \pm 2.00$ & $2.081 \pm 1.200$ \\
$4$&$2.7$   & $0.7$  & $2$ & $2.457 \pm 0.718$ & $96.79 \pm 1.94$ & $2.120 \pm 0.930$ \\
$5$&$2.7$   & $0.5$  & $4$  & $2.513 \pm 0.712$ & $96.34 \pm 1.96$ & $2.277 \pm 0.945$ \\
$6$&$2.7$   & $0.5$  & $8$  & $2.575 \pm 0.717$ & $95.24 \pm 2.72$ & $2.409 \pm 1.017$ \\
\rowcolor{gray!10}
7&\textbf{2.7} & \textbf{0.5} & \textbf{2} & 
$\mathbf{2.395 \pm 0.716}$ & 
$\mathbf{96.83 \pm 1.94}$ & 
$\mathbf{2.062 \pm 0.910}$ \\
\bottomrule
\end{tabular}
}
\label{tab3}
\end{table}

\subsubsection{Qualitative Evaluation}
\begin{figure*}
    \centering
    \includegraphics[width=\linewidth]{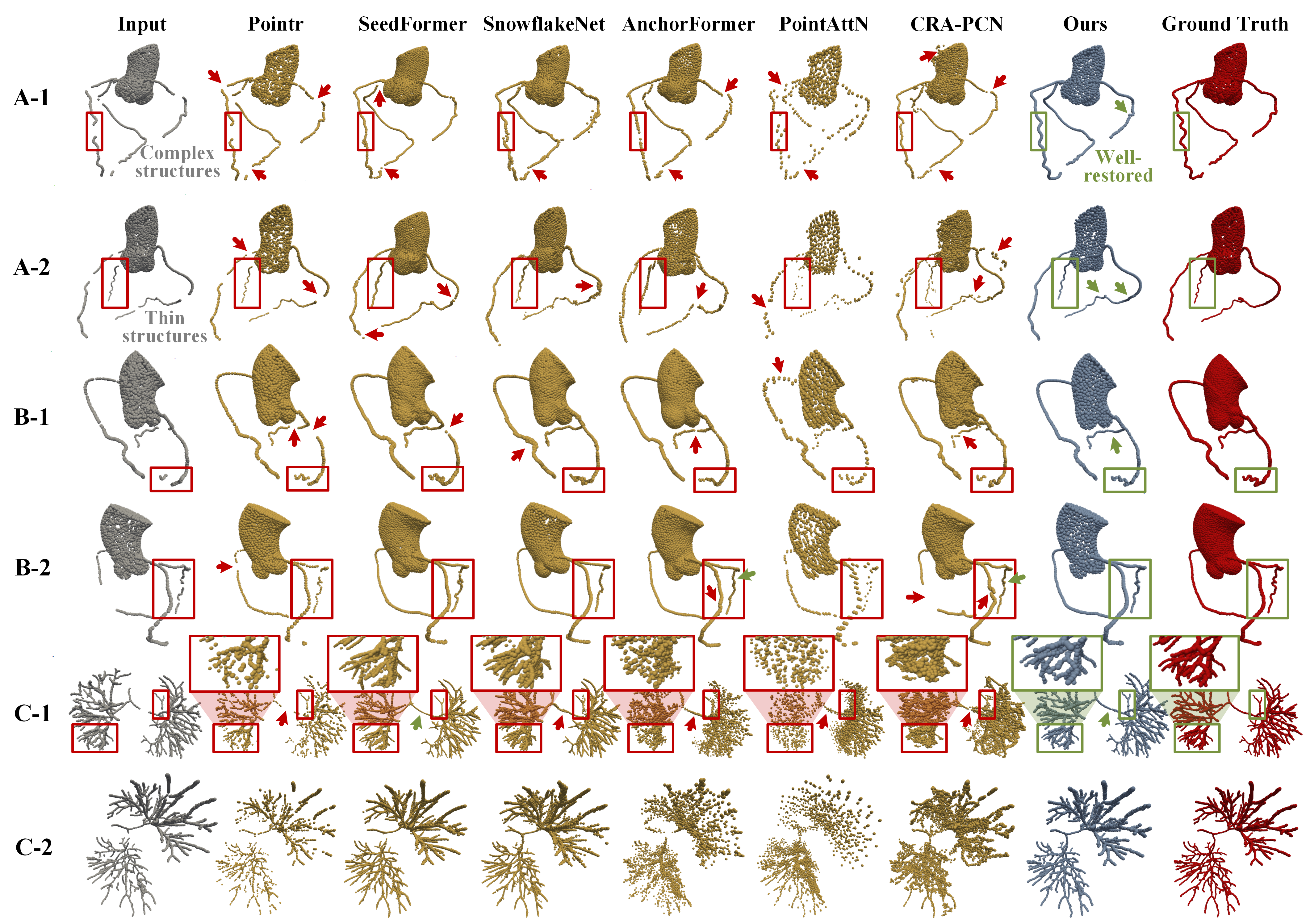}
    \caption{\textbf{Visual comparison of different point cloud-based models}. A-1, A-2 are two cases from our PC-CAC dataset. B-1, B-2 are two cases from PC-ImageCAS dataset. C-1, C-2 are two cases from PC-PTR dataset.}
    \label{fig3}
\end{figure*}

As shown in Fig.~\ref{fig3}, six representative cases from the PC-CAC, PC-ImageCAS, and PC-PTR datasets illustrate the superior completion performance of our \ourmodel{} compared to existing approaches. Our framework demonstrates better structural integrity, finer detail preservation, and improved robustness in reconstructing missing regions. 

\begin{figure*}
    \centering
    \includegraphics[width=\linewidth]{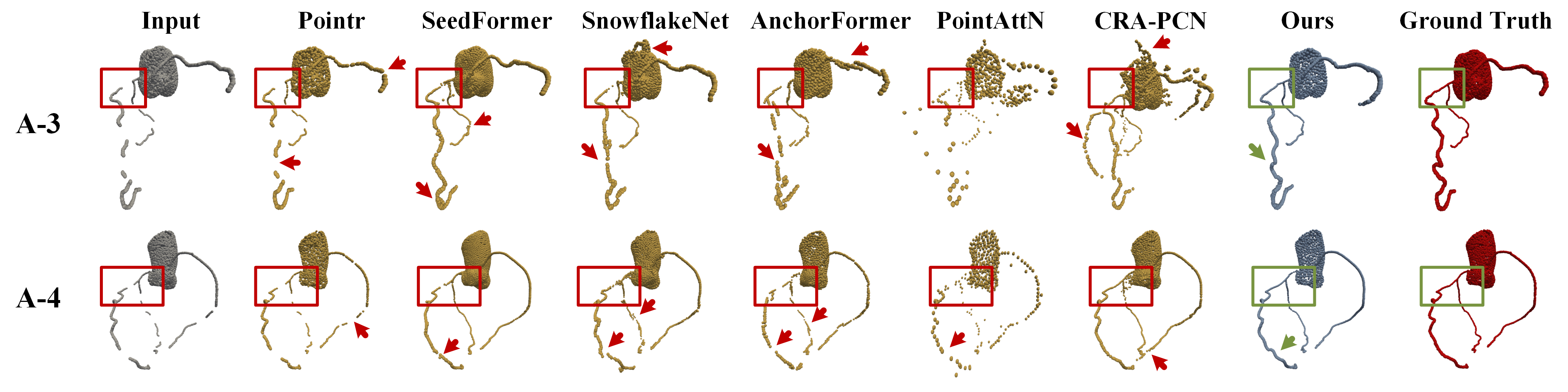}
    \caption{\textbf{Challenging cases with highly tortuous and complex anatomies}. A-3,4 are two cases from our PC-CAC dataset.}
    \label{fig:curve}
\end{figure*}

In our \textbf{PC-CAC} dataset (A-1, A-2), our method effectively restores complex and thin vascular structures, reducing discontinuities and achieving a more accurate topology compared to other methods. Referring to A-1 and A-2, it can be observed that PointAttN and CRA-PCN struggle to reconstruct and complete curved structures, leading to unintended fractures in originally intact regions. While Pointr performs relatively well in structural restoration, it fails to focus on elongated tubular structures, making completion infeasible. SeedFormer, SnowflakeNet, and AnchorFormer show comparatively better results. However, they still suffer from structural distortions, incomplete connections, and even incorrect reconstructions, particularly in complex or thin structures. In contrast, our method demonstrates superior robustness and performance, effectively preserving structural integrity while accurately reconstructing both complex and elongated tubular structures. Similarly, in the \textbf{PC-ImageCAS} dataset (B-1, B-2), our approach better captures the global shape while maintaining fine anatomical details, whereas other methods suffer from missing branches and distortions. For the \textbf{PC-PTR} dataset (C-1, C-2), which involves highly intricate pulmonary vessels, our model achieves remarkable consistency with the ground truth, accurately reconstructing fine-grained structures while avoiding excessive artifacts. In contrast, alternative methods exhibit structural collapse, missing connections, or distorted reconstructions, as highlighted by the red-marked error regions. These qualitative results further confirm the effectiveness of our detail-preserved feature extractor and multi-stage refinement strategy, enabling our model to accurately reconstruct complex tubular structures while minimizing artifacts and preserving structural coherence. 

To further investigate the performance under structurally extreme conditions, we highlight two particularly challenging cases with \textbf{\textit{highly tortuous and complex anatomies}} from our \textbf{PC-CAC} dataset (A-3, A-4). As shown in Fig.~\ref{fig:curve}, these examples are characterized by extensive curvature and disconnections, posing significant challenges to existing methods. It is evident that PointAttN and CRA-PCN struggle with reconstructing curved structures and may even introduce errors in previously intact regions. While Pointr shows a reasonable ability to restore structure, it lacks sensitivity to fine tubular details, leading to incomplete reconstructions. SeedFormer, SnowflakeNet, and AnchorFormer yield relatively better results but still exhibit structural distortions, missing connections, and sometimes incorrect linkages, particularly in complex and elongated vascular regions. In contrast, our method consistently achieves more accurate, well-connected, and robust reconstructions, more closely aligning with the ground truth.

\subsection{Compared with voxel-based methods}
\label{res:voxel}
To further evaluate the effectiveness of our \ourmodel{}, we compare it against voxel-based tubular structure completion methods. Specifically, we select two recent and representative baselines, DRTT and VSR-Net, which have demonstrated good performance in voxel-domain vascular reconstructions. Both quantitative and qualitative comparisons are conducted on the public PC-ImageCAS dataset for fairness.

\subsubsection{Quantitative Evaluation}
\textbf{Comparative Results.} The quantitative results are presented in Table~\ref{tab:voxel}. Our \ourmodel{} achieves the best performance across all three evaluation metrics. Specifically, our method obtains a $C\!D^{\ell_1}$ of 2.395, which is significantly lower than that of VSR-Net (2.529) and DRTT (2.869), indicating superior accuracy in structural alignment. In addition, \ourmodel{} achieves the highest F1-score of 96.83\% and the lowest Fidelity Error of 2.062, reflecting its ability to preserve both geometric detail and global continuity. These results demonstrate that operating directly in the point cloud domain provides advantages over voxel-based methods, particularly in terms of structural fidelity and precise reconnection.

\subsubsection{Qualitative Evaluation}
\begin{figure*}
    \centering
    \includegraphics[width=0.75\linewidth]{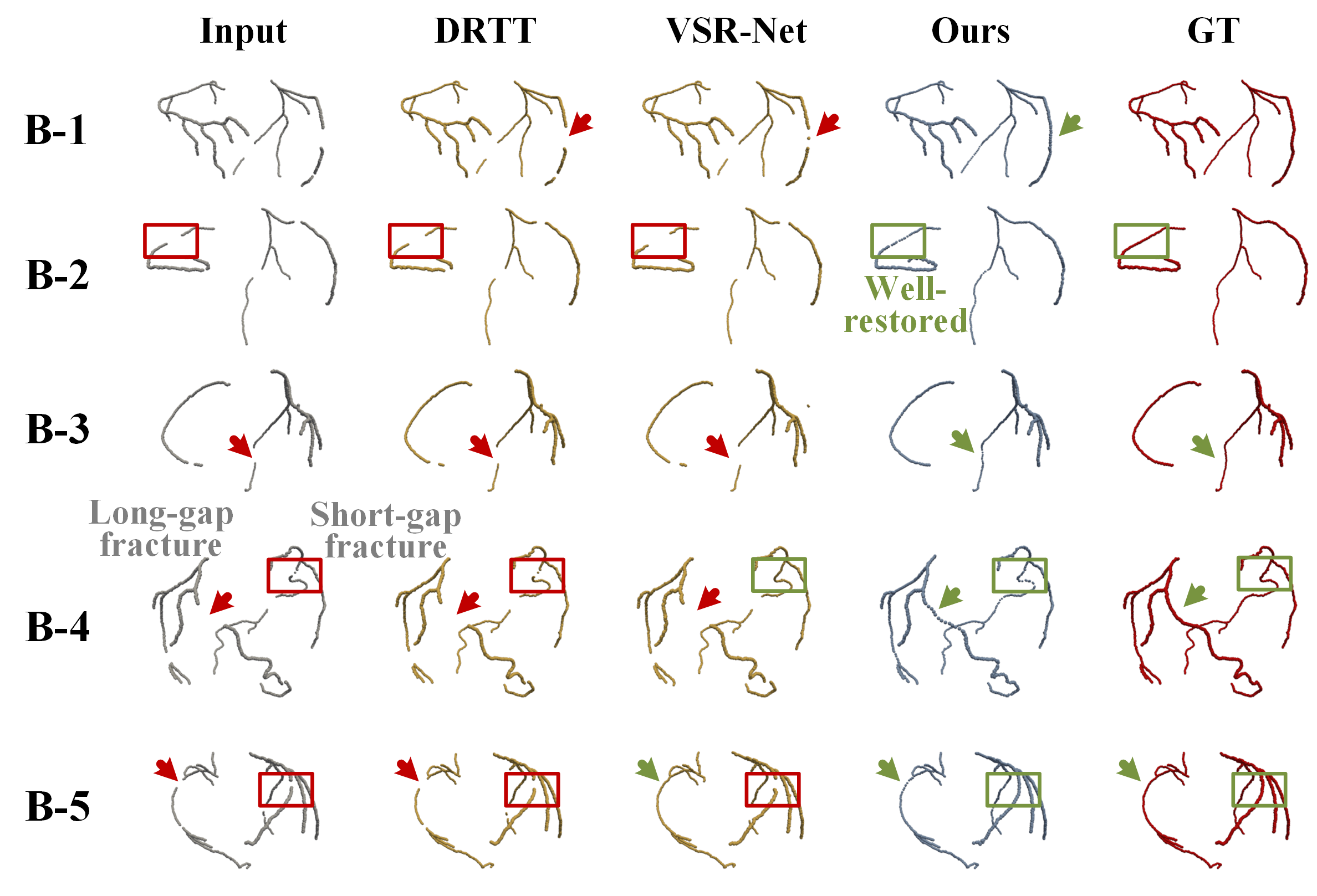}
    \caption{\textbf{Visual comparison of different voxel-based models}. B-1 to B-5 are 5 cases from the public PC-ImageCAS dataset. We focus on \textbf{\textit{vascular regions}} that are particularly difficult to reconnect. In particular, cases B-2 and B-4 involve long-gap structural discontinuities that pose significant challenges for reconstruction.}
    \label{fig:voxels}
\end{figure*}

Visual comparisons of different methods are presented in Fig.~\ref{fig:voxels}, where fice representative cases from the public PC-ImageCAS dataset are selected to evaluate completion performance under various structural challenges. We focus particularly on vascular regions that are difficult to reconnect. For instance, certain cases from PC-ImageCAS exhibit long-gap structural discontinuities that pose significant reconstruction difficulties for existing methods. Our \ourmodel{} consistently outperforms other approaches in restoring continuity and preserving anatomical fidelity across all datasets. The visual results demonstrate that our framework achieves better structural integrity, finer detail preservation, and improved robustness in reconstructing missing regions.

In the \textbf{PC-ImageCAS} dataset (B-1 to B-5), each row shows a different anatomical structure with varying types and severities of vascular disconnection. DRTT relies on a centerline template as input guidance. When this centerline contains fractures, the model fails to generate any meaningful reconnection in the corresponding regions, as evident in cases such as B-1 to B-5. As a result, large disconnected segments remain unreconstructed. In contrast, VSR-Net is capable of handling some short-range disconnections and successfully reconnects smaller fractures. However, it struggles in cases involving long-gap or high-curvature discontinuities (B-2 and B-4), where the restored structures remain incomplete or topologically inaccurate. Our method, by operating directly on the point cloud without requiring prior centerline input, demonstrates superior adaptability across all cases. In particular, in B-4, which includes both long-gap and short-gap fractures, our \ourmodel{} successfully restores both segments and achieves a structurally consistent result that closely matches the ground truth. The qualitative results highlight the robustness of our model in restoring complex vascular geometry under varying levels of structural loss.

\begin{table}[htb]
\label{tab:voxel}
\caption{Quantitative results of different voxel-based methods on public dataset PC-ImageCAS. \textbf{Bold} indicates the best results.}  
\centering
\resizebox{0.5\textwidth}{!}{
\begin{tabular}{c|c|c|c}
\toprule
&$CD^{\ell_1}(10^{-3})\downarrow$ &
F1(\%)$\uparrow$ &
Fidelity($10^{-3}$)$\downarrow$ \\
\midrule
DRTT~\citep{li2021deep}    & $2.869 \pm 0.630$ & $96.41 \pm 1.03$ & $3.765 \pm 0.627$ \\
VSR-Net~\citep{ye2025vsr}  & $2.529 \pm 0.719$ & $96.54 \pm 1.15$ & $3.205 \pm 0.910$ \\
\rowcolor{gray!10}
Ours & 
$\mathbf{2.395 \pm 0.716}$ & 
$\mathbf{96.83 \pm 1.94}$ & 
$\mathbf{2.062 \pm 0.910}$ \\
\bottomrule
\end{tabular}
}
\end{table}

\begin{figure}[!t]
    \centering
    \includegraphics[width=0.65\linewidth]{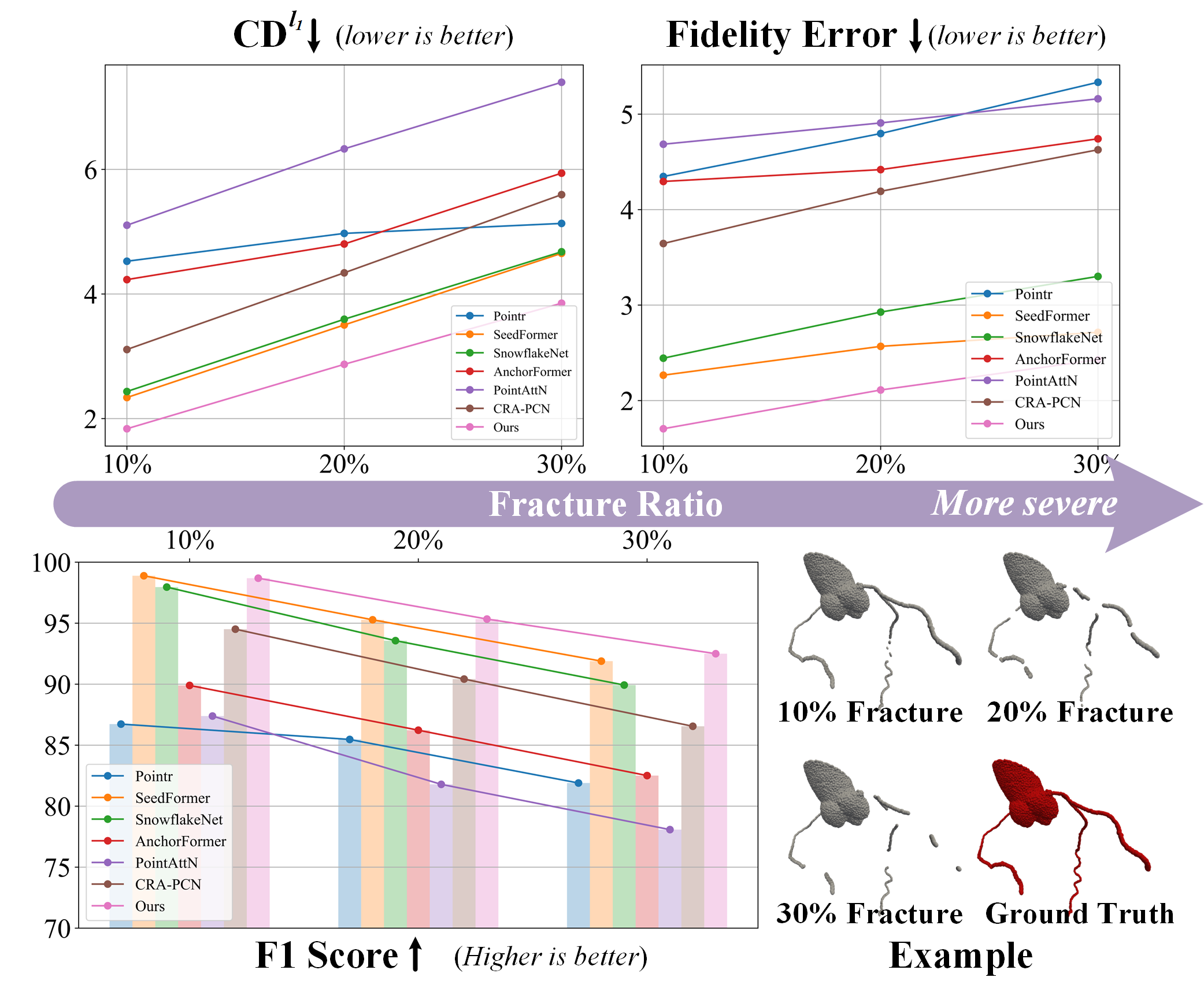}
    \caption{Performance comparison under varying fracture ratios (10\%, 20\%, and 30\%) in our PC-CAC dataset.}
    \label{fig: analysis}
\end{figure}

\subsection{Model Analysis – Completion capability under different fracture severity}
To further explore the completion limits of each model and objectively evaluate the reconnection performance of different methods under varying levels of structural disruption, we simulate fractures at different severities (10\%, 20\%, and 30\%) and compare the completion results across multiple baseline methods. The bottom-right section of Fig.~\ref{fig: analysis} illustrates a specific example simulating three different levels of fractures. At \textbf{10\%} fracture severity, a single fracture occurs in a complex and elongated region. At \textbf{20\%} fracture severity, multiple fractures emerge at critical locations such as thin structures and bifurcation points. At \textbf{30\%} fracture severity, a primary trunk is almost entirely fragmented, leaving only scattered points.
Both the line and bar chart in Fig.~\ref{fig: analysis} illustrate our model consistently outperforms others, achieving the lowest $C\!D^{\ell_1}$ and Fidelity Error, while maintaining the highest F1 across all fracture ratios. Notably, as the severity of fractures increases, competing methods exhibit a significant decline in performance, struggling with structural continuity and detail preservation. In contrast, our model remains robust, demonstrating superior completion accuracy and preserving fine structures even under severe fractures, highlighting its strong generalization capability in reconstructing complex tubular structures.

\subsection{Model Analysis – Impact of fracture type and severity}

\begin{figure}[htb]
    \centering
    \includegraphics[width=\linewidth]{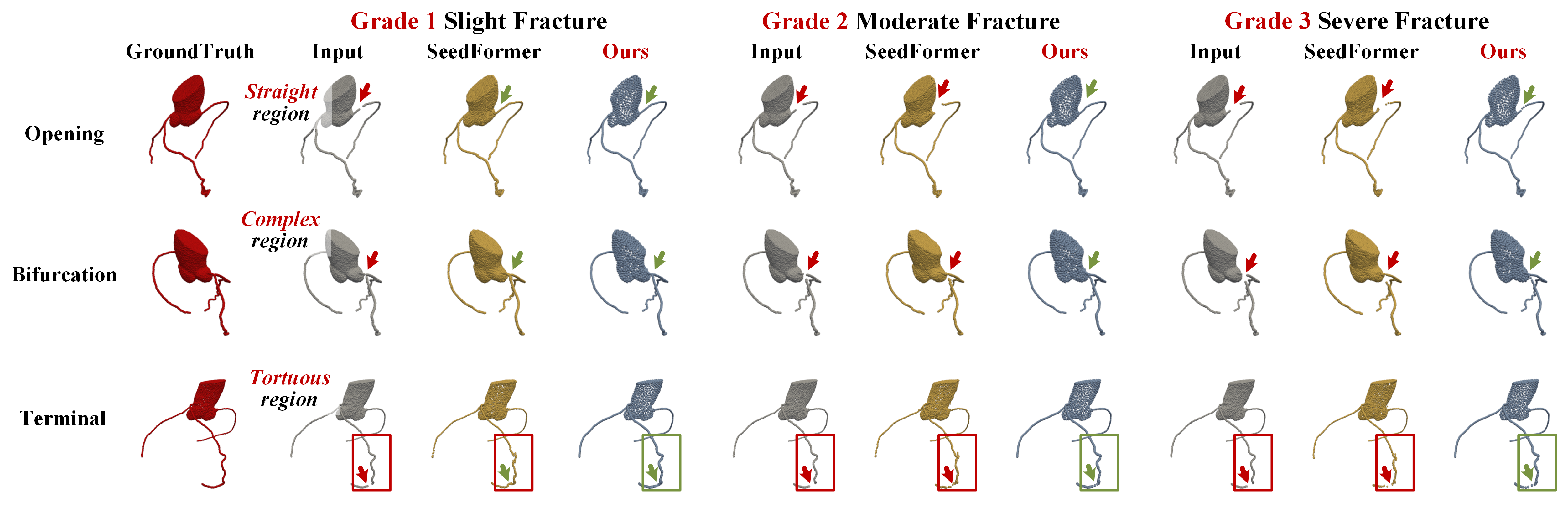}
    \caption{Visual comparison of reconnection results across different fracture severities and structural types. We compare the performance of SeedFormer (the second best) and our method under three grades of fracture severity: slight (Grade 1), moderate (Grade 2), and severe (Grade 3).}
    \label{fig: fracture}
\end{figure}

To further investigate the robustness of different models across diverse structural categories, we evaluate the reconnection performance on three representative types of anatomical regions: \textbf{\textit{opening}}, \textbf{\textit{bifurcation}}, and \textbf{\textit{terminal}} segments. As shown in Fig.~\ref{fig: fracture}, we simulate fractures with increasing severity (Grade 1–3), corresponding to slight, moderate, and severe levels of disconnection. Each row represents a distinct structural region, while each column pair compares the performance of SeedFormer (the second best model) and our \ourmodel{} under different fracture conditions. In \textbf{Grade 1 (Slight Fracture)}, all models generally perform well in reconnecting openings and bifurcation points, where disconnections are minor and structural cues remain clear. However, in terminal regions with highly tortuous geometries, SeedFormer often fail to maintain geometric fidelity, resulting in disrupted or overly smoothed reconstructions. In contrast, our \ourmodel{} preserves the integrity of these curved structures more effectively. In \textbf{Grade 2 (Moderate Fracture)}, the fractures predominantly occur around complex bifurcation regions, where multiple branches intersect at varying angles. SeedFormer typically reconnects only a single dominant branch, failing to capture the full bifurcation topology. In contrast, our method accurately reconstructs all intersecting branches, maintaining topological completeness and structural coherence across the bifurcation. In \textbf{Grade 3 (Severe Fracture)}, as the degree of disconnection increases, the reconstruction task becomes significantly more challenging due to the loss of large structural segments and spatial cues. Under such extreme conditions, SeedFormer suffers from unstable predictions, including fragmented outputs. In contrast, our method maintains stable and coherent reconstructions, consistently producing anatomically plausible results that align more closely with the ground truth.


\subsection{Future works}
Looking ahead, point cloud-based tubular structure completion offers a strong foundation for preserving anatomical integrity. Building upon this, integrating point cloud representations with corresponding voxel or image-based modalities holds great potential for further refining centerline estimation and improving structural accuracy. Such multi-modal fusion is expected to enhance model stability and provide more reliable reconstructions for downstream clinical applications such as diagnosis, simulation, or intervention planning. We will continue to release updates to the PC-CAC dataset, including the addition of paired image data, to promote ongoing research and foster broader advancements in this domain.

\section{Discussion and Conclusion}

Structural discontinuity remains a fundamental challenge in medical image analysis of tubular structures, severely affecting the accuracy of downstream tasks such as flow simulation and structural quantification. Existing methods often fail to preserve anatomical coherence when dealing with complex morphologies or long-gap fractures. To address this, we propose a dedicated tubular structure completion framework based on point cloud modeling, which significantly improves the capacity to restore fine-grained topology across different types of structural degradation. In addition, we release a novel dataset PC-CAC and standardized benchmark to facilitate reproducible research and fair comparison in tubular structure reconnection tasks. Experimental results across three datasets (PC-CAC, PC-ImageCAS, and PC-PTR) confirm the robustness and generalizability of our approach in handling diverse anatomical patterns. Our \ourmodel{} integrates a detail-preserved feature extractor, a progressive dense refinement strategy, and a global-to-local loss formulation, enabling accurate restoration of severely disconnected tubular structures while minimizing topological distortion. These findings underscore the importance of topology-aware modeling for medical imaging and suggest a promising direction for integrating structure-completion modules into broader diagnostic pipelines and patient-specific analysis systems.

In conclusion, this study introduces a novel approach for tubular structure reconnection from a unique point cloud perspective, addressing the limitations of traditional voxel-based methods in handling discontinuities within complex tubular structures. To our best knowledge, we construct the Point Cloud-based Coronary Artery Completion (PC-CAC) dataset, derived from real clinical data for the first time, providing a new benchmark for tubular structure reconnection tasks. Specifically, we propose the \ourmodel{}, a dedicated Tubular Structure Reconnection Network, which integrates a detail-preserved feature extractor, a multiple dense refinement strategy, and a global-to-local loss function. These components collectively enhance structural integrity, detail preservation, and the reconstruction of missing regions. Experiments on PC-CAC and two additional public datasets (PC-ImageCAS and PC-PTR) validate the effectiveness of our method, demonstrating state-of-the-art performance. Our method consistently achieves the lowest Chamfer Distance and Fidelity Error while maintaining the highest F1-score, confirming its superiority in restoring fragmented tubular structures with high precision. Additionally, our model remains robust even under high-severity fractures, proving its strong capability. Overall, this work lays a solid step toward structure-aware modeling of fractured anatomy in medical imaging, paving the way for more effective and equitable applications in clinical diagnosis.

\section*{Declaration of competing interest}
The authors declare that they have no known competing financial interests or personal relationships that could have
appeared to influence the work reported in this paper.


\printcredits

\bibliographystyle{cas-model2-names}

\bibliography{cas-refs}



\end{document}